\documentclass[letterpaper, 10 pt, conference]{ieeeconf}  % Comment this line out if you need a4paper
\usepackage{amsmath,amsfonts}
\usepackage[utf8]{inputenc}
\usepackage{graphicx}
\usepackage{float}
\usepackage{algpseudocode}
\usepackage[linesnumbered,ruled]{algorithm2e}
\usepackage{cite}
\usepackage{amsmath}
\usepackage{booktabs}
\usepackage{multirow}
\usepackage{subfigure}

\IEEEoverridecommandlockouts                              % This command is only needed if 
\begin{document}
\title{\LARGE \bf
Enhancing Multi-Agent Systems via Reinforcement Learning with LLM-based Planner and Graph-based Policy
}

% \author{\IEEEauthorblockN{
%    Ziqi Jia\IEEEauthorrefmark{1}\IEEEauthorrefmark{2},
%     Xiaoyang Qu\IEEEauthorrefmark{2},
%     Chenghao Liu\IEEEauthorrefmark{1},
%     and Jianzong Wang\IEEEauthorrefmark{2}}
%     \IEEEauthorblockA{\IEEEauthorrefmark{1}Tsinghua Shenzhen International Graduate School, Tsinghua University, Shenzhen, China}
%     \IEEEauthorblockA{\IEEEauthorrefmark{2}Ping An Technology (Shenzhen) Co., Ltd, Shenzhen, China}
%     % \IEEEauthorblockA{\IEEEauthorrefmark{3}The Shenzhen International Graduate School, Tsinghua University, Shenzhen, China}
% }

\author{Ziqi Jia$^{1,2}$, Junjie Li$^{1,3}$, Xiaoyang Qu$^{1}$, Jianzong Wang$^{1,*}$% <-this % stops a space
\thanks{$^{1}$ Ping An Technology (Shenzhen) Co., Ltd., Shenzhen, China}%
\thanks{$^{2}$ Tsinghua Shenzhen International Graduate School, Tsinghua University, Shenzhen, China
        % {\tt\small albert.author@papercept.net}
        }%
\thanks{$^{3}$ Huazhong University of Science and Technology, Wuhan, China
        % {\tt\small b.d.researcher@ieee.org}
        }%
\thanks{$^{*}$Jianzong Wang is the corresponding author, {\tt\small jzwang@188.com}}%
}

\maketitle

\thispagestyle{empty}
\pagestyle{empty}
% \author{\IEEEauthorblockN{Michael Shell}

% \maketitle
% \thispagestyle{empty}
% \pagestyle{empty}

%%%%%%%%%%%%%%%%%%%%%%%%%%%%%%%%%%%%%%%%%%%%%%%%%%%%%%%%%%%%%%%%%%%%%%%%%%%%%%%%
\begin{abstract}

Multi-agent systems (MAS) have shown great potential in executing complex tasks, but coordination and safety remain significant challenges. Multi-Agent Reinforcement Learning (MARL) offers a promising framework for agent collaboration, but it faces difficulties in handling complex tasks and designing reward functions. The introduction of Large Language Models (LLMs) has brought stronger reasoning and cognitive abilities to MAS, but existing LLM-based systems struggle to respond quickly and accurately in dynamic environments. To address these challenges, we propose LLM-based Graph Collaboration MARL (LGC-MARL), a framework that efficiently combines LLMs and MARL. This framework decomposes complex tasks into executable subtasks and achieves efficient collaboration among multiple agents through graph-based coordination. Specifically, LGC-MARL consists of two main components: an LLM planner and a graph-based collaboration meta policy. The LLM planner transforms complex task instructions into a series of executable subtasks, evaluates the rationality of these subtasks using a critic model, and generates an action dependency graph. The graph-based collaboration meta policy facilitates communication and collaboration among agents based on the action dependency graph, and adapts to new task environments through meta-learning. Experimental results on the AI2-THOR simulation platform demonstrate the superior performance and scalability of LGC-MARL in completing various complex tasks.

\end{abstract}

%%%%%%%%%%%%%%%%%%%%%%%%%%%%%%%%%%%%%%%%%%%%%%%%%%%%%%%%%%%%%%%%%%%%%%%%%%%%%%%%
\section{INTRODUCTION}

Recently, there has been a growing interest in multi-agent systems (MAS) due to their immense capability to handle various tasks \cite{venkata2023kt,liu2023robust,martin2023multi,chakraa2023optimization}. These systems are capable of handling simpler tasks like cleaning and surveillance, as well as more complex missions such as search and rescue missions or environmental surveillance \cite{chen2024scalable,zhang2022multi,chen2024meta}. However, coordinating multiple agents and ensuring safety—particularly in terms of collision avoidance \cite{yao2020multi}—remain major challenges. The complexity of these systems increases as the number of agents increases.

Multi-Agent Reinforcement Learning (MARL) provides a promising framework for agent collaboration. In this framework, agents learn optimal strategies by interacting with their environment, which enhances their ability to work together \cite{zhang2021multi,oroojlooy2023review}. For example, MAPPO \cite{kang2023cooperative} uses a centralized critic to facilitate coordination among agents, while QMIX \cite{rashid2020weighted} learns a global joint action value function by combining local Q-values to achieve multi-agent coordination. Despite the progress made, MARL still faces significant challenges when applied to real-world tasks. MARL struggles to execute complex human instructions and requires manual setting of task objectives as well as the design of intricate reward functions. These settings need to be adjusted as the scenarios change, which puts a heavy demand on the user's depth of understanding and experience with MARL.

The development of Large Language Models (LLMs) has introduced unprecedented reasoning and cognitive capabilities to robotics \cite{huang2022inner,liang2023code,singh2023progprompt,wu2023tidybot}. LLMs enhance the understanding of human intent and facilitate the prediction and execution of complex task sequences, improving the naturalness and effectiveness of human-machine interaction \cite{lin2023text2motion,ding2023task,DBLP:conf/corl/0003GFKLACEHHIX23,chen2024autotamp,irpan2022can}. In multi-agent systems, the application of LLMs has shown great potential by fostering communication and cooperation among agents, leading to more efficient collaborative workflows \cite{chen2024llm}. However, current LLM-based MAS often rely heavily on dialogues between LLMs \cite{mandi2024roco}, which presents challenges for resource-constrained small robots. Additionally, in dynamic environments that require quick responses, this reliance on extensive dialogue can hinder the system's scalability and efficiency.

\begin{figure*}[t!]
    \centering
    \includegraphics[width=18cm]{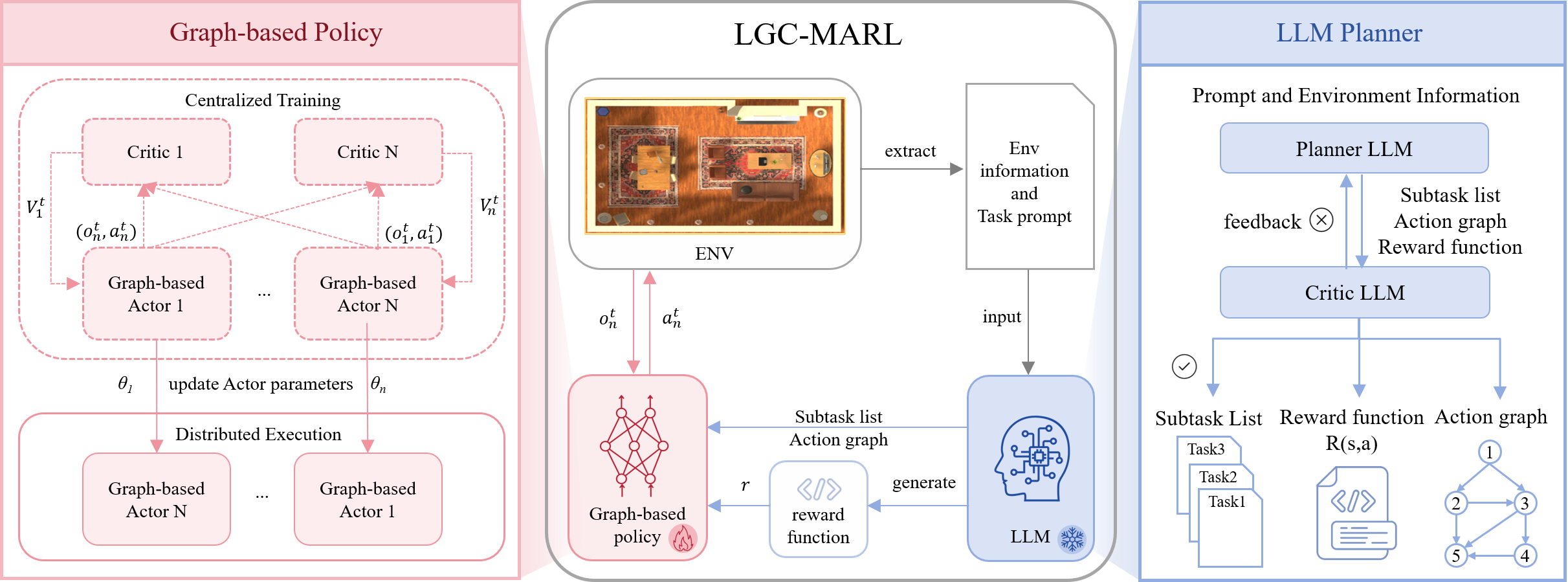}
    \caption{The illustration of the proposed framework. The sub-figure in the middle shows the LGC-MARL framework, with the two key modules showcasing the environment interactions and the prompt input, the bottom two parts highlighting the graph-based policy (in red) and the LLM planner (in blue).  When a new task is encountered, the two surrounding boxes reveal their internal workflows.}
    \label{fig:1}
\end{figure*}
% \vspace{-5cm} %调整图片与上文的垂直距离
To address the aforementioned challenges and effectively integrate  LLMs with MARL for multi-agent tasks, we propose the LLM-based Graph Collaboration MARL (LGC-MARL) framework. This framework leverages LLMs to decompose complex tasks into a series of low-level subtasks, and employs an action dependency graph to guide communication and cooperation among MARL agents. The underlying idea is that LLMs, equipped with rich world knowledge, can comprehend tasks and make plans based on the environment. By implicitly guiding the decision-making dependencies of MARL agents through the low-cost graph communication, LLMs enable MARL to iteratively learn optimal multi-agent collaboration policies through interaction with the environment. We introduce a critic-equipped LLM planner to extract environmental and task information, generate a sequence of subtasks, and employ another critic LLM module to verify the correctness of these subtasks, mitigating the hallucination problem of LLMs. To address the over-reliance on dialogues when using existing LLM data, we constrain the LLM's output to a graph format and adopt a graph-based collaboration meta policy for MARL agents to coordinate their actions, enabling flexible and efficient handling of LLM instructions. To tackle the challenging problem of designing reward functions in MARL, we propose an LLM-based reward function generator and carefully design prompts to guide the LLM to consider the problem from a MAS perspective, thus generating reward functions that promote agents collaboration.
%策略的名字再想一下
% \begin{figure*}[t!]
%     \centering
%     \includegraphics[width=18cm]{Fig1.png}
%     \caption{framework}
%     \label{fig:galaxy}
% \end{figure*}
Our main contributions include:
\begin{itemize}
    \item We have developed the LGC-MARL framework, integrating LLMs with MARL. By leveraging LLMs to decompose complex tasks into subtasks and employing action dependency graphs to guide inter-agent communication and cooperation, our method enables MARL agents to iteratively learn optimal collaborative strategies. Furthermore, by benefiting from the rich world knowledge and planning capabilities of LLMs, our approach enhances the efficiency of multi-agent tasks.
    % \item We developed the LGC-MARL framework, integrating large language models with MARL. Specifically, the large language model breaks down tasks into sub-tasks and evaluates them through a critic model. It guides MARL collaboration using decision dependency graphs and generates dense reward functions.
    \item We introduce a graph-based collaboration meta policy and an LLM-based reward function generator for MARL. Therefore, our agents can collaborate efficiently guided by an action dependency graph, and rapidly adapt to new task environments through meta-learning. Furthermore, our LLM-based reward generator tailors reward functions to the specific dynamics of MAS, enabling efficient learning and collaboration in MARL.
    \item  Compared to existing methods, LGC-MARL exhibits significant advantages in terms of task complexity, adaptability, and collaborative efficiency.
\end{itemize}

% \begin{figure*}[t!]
%     \centering
%     \includegraphics[width=18cm]{Fig1.png}
%     \caption{framework}
%     \label{fig:galaxy}
% \end{figure*}

\section{RELATED WORKS}

\subsection{LLM for Multi-agent Systems}
DyLAN\cite{liu2023dynamic} introduces a dynamic adjustment mechanism, allowing LLM agents to adapt their interaction modes based on real-time performance and task demands. EMMA\cite{DBLP:conf/cvpr/Yang0LTLS00S24} advances this concept by using centralized evaluation and natural language communication to fine-tune LLMs online, simplifying coordination among multiple agents. Chen et al.\cite{chen2023multi} studied how LLMs help multi-agent systems reach consensus, examining the impact of agent attributes, numbers, and network topology on the consensus process. CoELA \cite{zhang2023building} offers a modular approach, integrating LLMs into perception, memory, communication, and planning modules, facilitating collaborative task completion among agents. SMART-LLM \cite{DBLP:conf/iros/KannanVM24} breaks down complex tasks into sequential stages, generating actions needed for complex goals. RoCo \cite{mandi2024roco} equips each robotic arm with LLMs, enabling them to infer task strategies and generate trajectory points through dialogue. Most existing studies rely on LLM communication, limiting efficiency in dynamic environments. They also lack exploration, hindering optimal solution discovery.
In contrast, our approach combines LLMs with MARL using graph structures, reducing overhead and improving decision quality for more resilient, exploratory multi-agent systems.
\subsection{LLM for Reinforcement Learning}
LLaRP \cite{DBLP:conf/iclr/SzotSAMMTMHT24} utilizes LLM to translate natural language instructions into task-specific unique languages and adopts an inside-out approach, avoiding direct exposure of RL strategies to natural language, thus promoting efficient learning. STARLING \cite{wu2024spring} leverages the LLM to translate natural language instructions into environment information and metadata, generating large amounts of text for RL agents to acquire necessary skills. In contrast, GLAM \cite{carta2023grounding} employs the LLM as a policy to select the next action, interacting directly with the environment and training through online reinforcement learning. Plan4MC \cite{baai2023plan4mc} acts as a high-level planner, utilizing the LLM to generate high-level action plans based on historical records and current environmental states, while combining independent low-level RL policies to execute specific actions. Eureka \cite{ma2023eureka} proposes a self-reflective reward optimization algorithm that samples candidate reward functions from an encoded LLM and evaluates these functions through RL simulation, gradually improving the reward output. Meanwhile, Text2Reward \cite{xie2023text2reward} formulates adaptable dense reward functions as executable programs derived from environment descriptions. It refines rewards by implementing learning strategies within the environment and incorporating human feedback.

\section{METHOD}
Our proposed framework is shown in Fig. \ref{fig:1}, with the following process: First, the LLM generates a sequence of plans and a dependency graph based on environmental information. These outputs are then validated and refined by a critic model to ensure their feasibility and effectiveness. Once approved, the plan sequences and dependency graph are passed to a MARL policy. The MARL policy interacts with the environment and receives rewards from a reward function generated by the LLM. This allows the agents to optimize their decisions and improve performance over time.

\subsection{Problem Setup}

% We model the scenario as a Multi-Agent Markov Decision Process (MMDP) $(S, \{A^i\}_{i=1}^{N}, \mathbb{P}, R, \gamma)$, with $s$ as the global state set, $A^i$ as agent $i$'s action space, and  $\boldsymbol{a} = (a^1, a^2, \ldots, a^N)$ for all agents joint action. $\mathbb{P}$ defines transition probabilities between states, while $R$ assigns rewards to state-action pairs. $\gamma$ discounts future rewards.
% We model the scenario as a Multi-Agent Markov Decision Process (MMDP) $(S, \{A^i\}_{i=1}^{N}, \mathbb{P}, R, \gamma)$, where $S$ is the set of global states, $A^i$ represents the action space of agent $i$, and we define $\boldsymbol{a} = (a^1, \dots, a^N)$ as the joint action of all $N$ agents. The transition probability function between states is defined by $\mathbb{P}$, while $R$ represents the reward function, and $\gamma$ is the discount factor.
We model the scenario as a Multi-Agent Markov Decision Process (MMDP) denoted by $(S, \{A^i\}_{i = 1}^{N}, \mathbb{P}, R, \gamma)$. Here, $S$ represents the set of global states. For each agent $i$, $A^i$ stands for its action space. We define the joint action of all $N$ agents as $\boldsymbol{a}=(a^1,\ldots,a^N)$. The state transition probability function is given by $\mathbb{P}$. Meanwhile, $R$ represents the reward function, and $\gamma$ is the discount factor.

Our objective is finding a joint policy $\pi^*$ maximizing the expected cumulative discounted reward:
\begin{equation}
\pi^* = \arg\max_\pi \mathbb{E}_{\pi}\left[\sum_{t=0}^{\infty} \gamma^t r_{t+1} \right],
\end{equation}
where $\mathbb{E}$ represents taking the expectation and $t$ denotes the time step.

\subsection{LLM Planner and Critic Model}
We adopt a pre-trained LLM as a planner, which generates a series of planning steps based on environmental observations and task descriptions. The generated planning sequence can be represented as a list $[\zeta_l]$, where \(l = 1, \ldots, L\), representing specific actions or decisions at each step. Additionally, the LLM constructs a graph structure that describes the decision dependencies among agents based on the generated planning sequence and the current positions of the agents. This graph not only reflects the collaborative relationships between different agents but also aids in further optimizing the effectiveness and coordination of decisions.
\begin{figure}[t!]
    \centering
    \includegraphics[width=8cm]{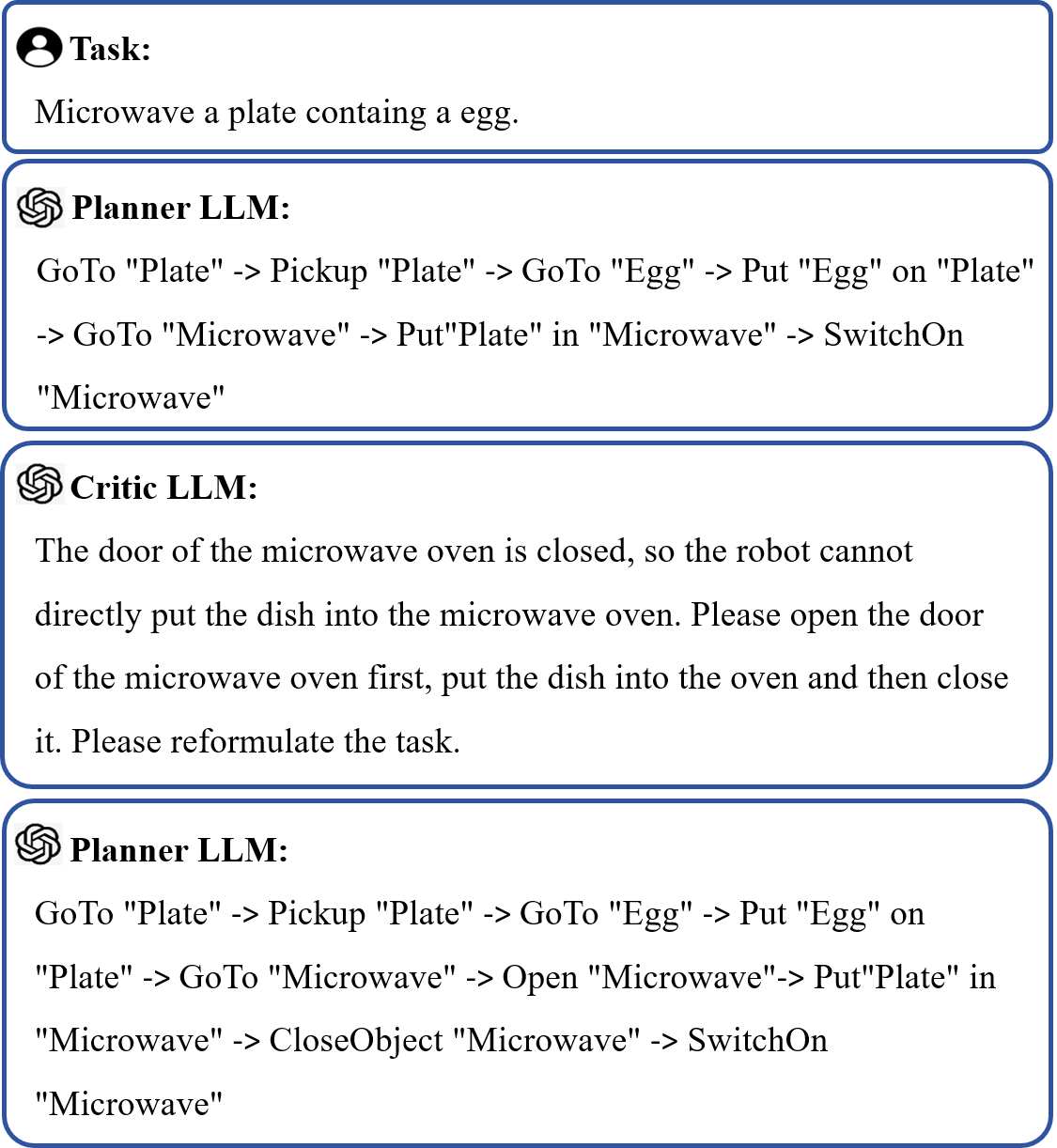}
    \caption{Planner LLM and Critic LLM}
    \label{fig:2}
\end{figure}
% \vspace{-0.2cm} %调整图片与上文的垂直距离
To ensure the executability of the generated planning sequence in practical operations, we introduce another LLM as a critic model to evaluate the plans produced by the planner. As shown in Fig. \ref{fig:2}, when the critic model identifies factual errors in the planning sequence, it conveys error information back to the planner through a feedback mechanism. Upon receiving this feedback, the planner revises the initial plan, generating an improved planning sequence and an updated decision dependency graph. Through the interaction between the planner and the critic model, we achieve continuous optimization of the plans, ensuring their effectiveness and accuracy during execution.
\subsection{LLM-based Reward Function Generator}
The setting of the reward function is a traditional challenge in reinforcement learning. Compared to single-agent reinforcement learning, multi-agent reinforcement learning requires consideration of the interests of all agents when designing the reward function. Additionally, due to the non-stationary nature of the environment from each agent's perspective, crafting effective MARL reward functions is a formidable task that involves extensive debugging.
\begin{figure}[t!]
    \centering
    \includegraphics[width=8cm]{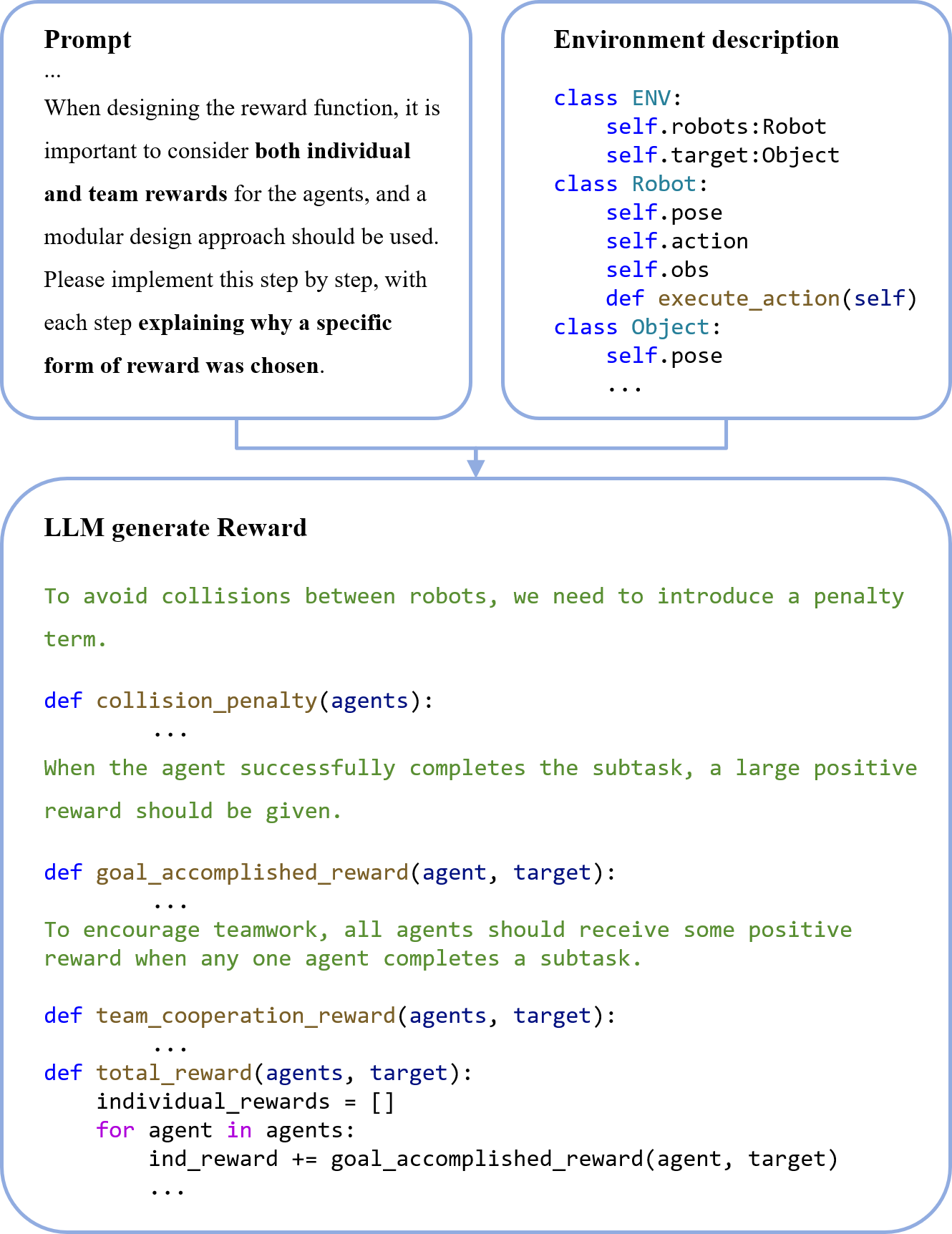}
    \caption{LLM-based reward function generator}
    \label{fig:3}
\end{figure}
As shown in Fig. \ref{fig:3}, we input the abstracted environmental descriptions and task requirements into the LLM. The LLM is instructed to generate a reward function that considers both individual and team rewards for the agents, with a modular design. Although the LLM can help create complex reward functions, maintaining interpretability is crucial. Therefore, instead of directly providing the final reward function, we require the LLM to demonstrate its reasoning process. This is done by generating a series of intermediate steps, with each step explaining the rationale behind choosing a particular form of reward.
% 改进后的
\begin{algorithm}

\SetAlgoLined

\DontPrintSemicolon

\caption{LGC-MARL.}

\label{alg:GCS_optimization_m}

Initialize $LLM$ and  graph-based collaboration meta policy $\pi^i_m$;

Initialize $L$, $C$, $B$ and task set $T$;

Initialize the policy network parameters $\theta^i$, $\phi^i$, where $i = 1, ..., N$;

Initialize the target networks $\theta^{i'} = \theta^i$ and $\phi^{i'} = \phi^i$;

Initialize data buffer $\mathcal{D} = \{\}$;

\For{each episode}{

    Get subtask list $[\zeta_l]$ from $LLM$;

    Get reward function $R$ from $LLM$;

    Initialize $h_{0,\pi}^{(1)}, \dots, h_{0,\pi}^{(N)}$ actor RNN states;

    Initialize $h_{0,V}^{(1)}, \dots, h_{0,V}^{(N)}$ critic RNN states;

    \For{each timestep $t$}{

        $\boldsymbol{G}_t = LLM(s_t)$;

        Get the adjacency matrix $\boldsymbol{M}_t = f_m(\boldsymbol{G}_t)$;

        \For{$i$ in order $\boldsymbol{M}_t$}{

            $a_t^i, h_{t,\pi}^{(i)} = \pi^i_m(o_t^i, h_{t-1,\pi}^i, a_t^{pt(i)}; T)$;

            $v_t^i, h_{t,V}^{(i)} = Q_t^{\pi^i_m}(s_t^i, h_{t-1,V}^i; T)$;

            Receive reward $r_t$ from $R$ ;

            Observe next state $o_{t+1}$;

            Add transition $\{o_t, a_t, r_t, o_{t+1}\}$ into $\mathcal{D}$;

        }

    }

    \If{$episodes > L$}{

        Sample a batch 

        $\mathcal{B} = \{o_j, a_j, r_j, o_{j+1}\}_{j=0}^B \sim \mathcal{D}$;

        Update $\pi^i_m$ using $\mathcal{B}$ and Eq. (5);

        Update $\phi^i$ using Eq. (6);

        Every $C$ steps reset $\theta^{i'} = \theta^i$ and $\phi^{i'} = \phi^i$

        for $i = 1, ..., N$;

    }

}
\end{algorithm}

\subsection{Graph-based Collaboration Meta Policy with LLM-based Graph Generator}

LLMs possess strong capabilities for text understanding and generation, effectively capturing environmental states. We use an LLM to translate the state of the environment and the behavior of agents into textual form, then employ the LLM to generate text that describes decision dependencies. This leads to the construction of a directed acyclic graph (DAG) $\boldsymbol{G}$, representing the dynamic decision-making structure among agents. Finally, we convert DAG $\boldsymbol{G}$ into an adjacency matrix $\boldsymbol{M}$. The formula is as follows:
\begin{equation}\label{eq:1}
\boldsymbol{M} = f(LLM(\left\{o^i\right\}_{i=1}^N)),
\end{equation}
where $\left\{o^i\right\}_{i=1}^N$ denotes the environment observed by all agents, and $f$ represents the function that converts the graph into an adjacency matrix.

To enable agents to quickly adapt to new environments and task scenarios, we incorporate meta-learning mechanisms into the graph-based reinforcement learning strategy. We define a set of tasks \( T \), and the graph-based collaboration meta policy can be expressed as:
\begin{equation}
    \pi_{m} = \prod_{i=1}^N \pi^i_{m}(a^i \mid o^i, a^{pt(i) \sim M}; T),
\end{equation}
% where  $\pi^i_{m}$ is the meta-learning policy of the \( i \)-th agent, which takes into account the information from the task set TT
% We can denote the decision graph-based coordinated policy as follows:
% \begin{equation}\label{eq:2}
% \pi=\prod_{i=1}^N \pi^i\left(a^i \mid o^i, a^{pt(i) \sim M}\right),
% \end{equation}
where $\pi^i_{m}$ is the meta-learning policy of the \( i \)-th agent, which takes into account the information from the task set \( T \), $o^i$ represents the observation of the \( i \)-th agent, $pt(i)$ denotes the parent agents of agent $i$, and $a^{pt(i) \sim M}$ are the actions taken by these parent agents, with their order determined by the LLM. Our overall objective is to maximize the cumulative revenue, which can be expressed as:
\begin{equation}\label{eq:3}
\eta_{m}=\mathbb{E}_{\boldsymbol{a} \sim \pi(\cdot \mid s, M)}\left[\sum_{k=0}^{\infty} \gamma^k r\left(s_{t+k}, \boldsymbol{a}_{t+k}\right)\right].
\end{equation}

% \begin{equation}\label{eq:3}
% \eta_{m}=\mathbb{E}_{A \sim \rho, \boldsymbol{a} \sim \pi(\cdot \mid s, A)}\left[\sum_{k=0}^{\infty} \gamma^k r\left(s_{t+k}, \boldsymbol{a}_{t+k}\right)\right].
% \end{equation}

In a multi-agent environment, each agent \( i \) (where \( i = 1, 2, \ldots, N \)) adheres to a specific policy \( \pi_m^i \). Collectively, these policies form the coordination strategy set of the entire system \( \Pi = \{\pi_m^1, \pi_m^2, \ldots, \pi_m^N\} \). These policies are parameterized by a set of parameters \( \theta = \{\theta^1, \theta^2, \ldots, \theta^N\} \). The gradient of expected return for each agent \( i \) can be written as follows:
% \begin{equation}
% \begin{aligned}
% &\nabla_{\theta^i} \eta_{m}(\theta^i) = \nabla_{\theta^i} \eta^i_{m} = \\
%     &\mathbb{E}_{a^i \sim \pi^i_{m}} \left[ \nabla_{\theta^i} \log \pi^i_{m}(a^i \mid o^i, a^{p t(i)}; T) \sum \rho(M \mid s) Q_{\pi_{m}}^i(s, a) \right],
% \end{aligned} 
% \end{equation}
\begin{equation}
\begin{split}
&\nabla_{\theta^i} \eta_{m}(\theta^i) = \nabla_{\theta^i} \eta^i_{m} \\
&= \mathbb{E}_{a^i \sim \pi^i_{m}} \left[ \nabla_{\theta^i} \log \pi^i_{m}(a^i \mid o^i, a^{p t(i)}; T) Q_{\pi_{m}}^i(s, a) \right],
\end{split}
\end{equation}

where $Q_{\pi_{m}}^i$ is the centralized action-value function, which can be updated to achieve iterative improvement as follows:
\begin{equation}
    L(\phi) = \mathbb{E}_{(s, \boldsymbol{o}, M, \boldsymbol{a}) \sim \mathcal{D}} \left[ \left( Q_{\pi_{m}}^i(s, \boldsymbol{a}) - y^i_{m} \right)^2 \right],
\end{equation}
\begin{equation}
    y_m^{i} = r + \gamma \max_{u'} Q_{{\pi_{m}^-}}^i(s', a'),
\end{equation}
% \begin{equation}\label{eq:4}
% \begin{aligned}
% &\eta^i=\mathbb{E}_{\boldsymbol{a} \sim \pi}\left[\pi(\boldsymbol{a}, A \mid s) Q_{\boldsymbol{\pi}}^i(s, \boldsymbol{u})\right] \\
% &=\mathbb{E}_{a^i \sim \pi^i}\left[\rho(M\mid s) \prod_{i=1}^N \pi^i\left(a^i \mid o^i, a^{pt(i) \sim M}\right) Q_{\boldsymbol{\pi}}^i(s, \boldsymbol{a})\right].
% \end{aligned}
% \end{equation}

% Accordingly, for each agent \( i \), the gradient of the expected utility function can be expressed in terms of the adopted policy parameters:
% \begin{equation}\label{eq:5}
% \begin{aligned}
% & \nabla_{\theta^i} \eta\left(\theta^i\right)=\nabla_{\theta^i} \eta^i \\
% & =\mathbb{E}_{a^i \sim \pi^i}\left[\nabla_{\theta^i} \log \pi^i\left(a^i \mid o^i, a^{p t(i)}\right)\sum \rho(M\mid s) Q_\pi^i(s, a)\right],
% \end{aligned}
% \end{equation}
% where $Q_{\pi}^{i}$ is the centralized action-value function, which can be updated to achieve iterative improvement as follows:
% \begin{equation}\label{eq:7}
% \mathcal{L}(\phi)=E_{(s, \boldsymbol{o}, A, \boldsymbol{a}) \sim \mathcal{D}}\left[\left(Q_{\boldsymbol{\pi}}^i(s, \boldsymbol{a})-y^i\right)^2\right],
% \end{equation}
where $y_m^{i}$ is the learning target with meta-learning features, $Q_{{\pi_{m}^-}}^i$ is the target network parameterized by \( \theta_-^i \), and \( \mathcal{D} \) is the experience replay buffer which records the experiences. The pseudocode is shown in \textbf{Algorithm \ref{alg:GCS_optimization_m}}.

\section{experiment}

\subsection{Experiment Setup}

% AI2-THOR\cite{kolve2017ai2} developed by the Allen Institute for AI is a simulation environment for indoor spaces, designed to support research for home and service robots. It provides a highly realistic 3D environment filled with furniture and everyday objects. Within this simulator, robots can perform a variety of actions such as 'MoveLeft' (moving left), 'MoveAhead' (moving forward), as well as object interactions like 'PickupObject' (picking up an object) and 'PutObject' (placing an object). We chose 4 scenes from AI2-THOR to test on and designed a set of tasks for each scene. And all LLMs used in the experiments are GPT-4.
AI2-THOR\cite{kolve2017ai2} is a 3D simulation environment for home robots. Within this simulator, Robots can perform actions such as \text{‘}MoveLeft\text{'}, \text{‘}MoveAhead\text{'}, \text{‘}PickupObject\text{'}, and \text{‘}PutObject\text{'}. We selected 4 scenes from AI2-THOR and designed specific tasks for each. All LLMs used were GPT-4.

\subsection{Evaluation metrics}

To comprehensively evaluate the performance of each method, we selected three key metrics: Success Rate (SR), Average Completion Time (AT), and Normalized Token Cost (TC).
\begin{itemize}
    \item \textbf{Success Rate}: A task is considered failed if there are collisions, interferences between agents, or if the task times out. By calculating the percentage of successfully completed tasks, we can directly assess the robustness and stability of each method.
    \item \textbf{Average Completion Time}: This metric measures the average time(in seconds) taken by agents to complete a task. A shorter time indicates higher efficiency of the agent and indirectly reflects the optimization level of the algorithm.
    \item \textbf{Normalized Token Cost}: To better compare the language model resource consumption of different methods, we normalized the token consumption \cite{chen2024scalable}. The specific calculation formula is as follows:
    \begin{equation}
        \hat{w}_i = \frac{w_i}{min(W)},
    \end{equation}
    where $W$ is the set of token costs, \( i \)-th represents the token cost of the \( i \)-th method, $\hat{w}_i$ is the normalized metric value, and the best possible value for $\hat{w}_i$ is 1.0. 
\end{itemize}

\begin{table*}[]
\centering
\caption{EVALUATION RESULTS ON FOUR SCENES. We reported the average success rate (↑) for each scene with a agent number of 4 over more than 100 runs, the average time (↓) per scene, and the normalized average token cost (↓).}
\begin{tabular}{@{}clcccccc@{}}
\toprule
\multicolumn{2}{c}{} & Centralization & LLMs Dialog & LGC-MARL & LGC-MARL & LGC-MARL & \multirow{2}{*}{LGC-MARL} \\
\multicolumn{1}{l}{} & \multicolumn{1}{l}{} & LLM & Dialog & (wo critic) & (wo reward) & (wo graph) &  \\ \midrule
\multirow{3}{*}{Scene1} & Success Rate & 0.6 & 0.68 & 0.76 & 0.89 & 0.73 & \textbf{0.92} \\
 & Average Time & 99.59 $\pm$ 5.32 & 154.54 $\pm$ 9.74 & 68.61 $\pm$ 2.09 & 73.15 $\pm$ 3.39 & 78.95 $\pm$ 3.96 & \textbf{66.47 $\pm$ 1.82} \\
 & Token Cost & 6.3 & 10.7 & 1.3 & \textbf{1} & 2.2 & 1.80 \\ \midrule
\multirow{3}{*}{Scene2} & Success Rate & 0.54 & 0.61 & 0.7 & 0.84 & 0.67 & \textbf{0.88} \\
 & Average Time & 107.74 $\pm$ 6.07 & 189.21 $\pm$ 11.82 & 80.94 $\pm$ 2.57 & 84.70 $\pm$ 3.46 & 88.24 $\pm$ 4.02 & \textbf{78.76 $\pm$ 2.42} \\
 & Token Cost & 7.8 & 14.2 & 2.6 & 2.3 & 3 & \textbf{2.1} \\ \midrule
\multirow{3}{*}{Scene3} & Success Rate & 0.57 & 0.64 & 0.74 & 0.86 & 0.69 & \textbf{0.90} \\
 & Average Time & 105.11 $\pm$ 5.83 & 182.31 $\pm$ 10.57 & 77.72 $\pm$ 2.63 & 80.08 $\pm$ 3.58 & 86.01 $\pm$ 3.35 & \textbf{75.93 $\pm$ 2.23} \\
 & Token Cost & 7.5 & 13.6 & 2.4 & \textbf{2.1} & 3.4 & 2.6 \\ \midrule
\multirow{3}{*}{Scene4} & Success Rate & 0.51 & 0.60 & 0.71 & 0.8 & 0.65 & \textbf{0.87} \\
 & Average Time & 115.85 $\pm$ 6.56 & 205.69 $\pm$ 11.31 & \textbf{85.75 $\pm$ 2.81} & 89.31 $\pm$ 4.24 & 95.37 $\pm$ 3.50 & \textbf{84.81 $\pm$ 2.65} \\
 & Token Cost & 8.8 & 16.3 & 4.2 & 3.8 & 5.1 & \textbf{3.5} \\ \bottomrule
\end{tabular}
% \caption{}
\label{tab:my-table1}
\end{table*}

\subsection{Results and Analysis}
\textbf{Comparison of Different LLM Methods.} As shown in Table \ref{tab:my-table1}, we conducted a comprehensive comparison between LGC-MARL and two prevalent LLM-based multi-agent methods: centralized LLM \cite{DBLP:conf/iros/KannanVM24} and LLM dialogue \cite{mandi2024roco}. 
Centralized LLM employs a single LLM to plan actions for all agents, limiting inter-agent collaboration and resulting in the lowest task success rate due to its inability to account for individual agent perspectives and dynamic environmental changes. In contrast, the LLM dialogue approach equips each agent with an independent LLM to facilitate coordination through dialogue, which enhances collaboration by enabling agents to exchange information and negotiate strategies. However, this approach also leads to the longest task completion time due to the computational overhead of natural language processing and the excessive token consumption from multi-round dialogues, which can become particularly problematic in complex scenarios. Furthermore, the increased number of dialogue rounds can exacerbate the risk of LLM hallucinations, as each interaction introduces opportunities for misinterpretation or generation of inconsistent information, further reducing task success. In comparison, our proposed LGC-MARL method demonstrated superior performance across all metrics. Whether it's task success rate, average completion time, or token consumption, our method consistently outperformed the baselines by leveraging structured communication and efficient decision-making mechanisms. This indicates that LGC-MARL effectively balances collaboration, efficiency, and resource utilization, making it a robust solution for multi-agent systems in diverse applications.

% Centralized LLM employs a single LLM to plan actions for all agents, limiting inter-agent collaboration and resulting in the lowest task success rate. In contrast, the LLM dialogue approach equips each agent with an independent LLM to facilitate coordination through dialogue. While this enhances collaboration, it also leads to the longest task completion time due to the computational overhead of natural language processing and the excessive token consumption from multi-round dialogues. Furthermore, the increased number of dialogue rounds can exacerbate the risk of LLM hallucinations, further reducing task success. In comparison, our proposed LGC-MARL method demonstrated superior performance across all metrics. Whether it's task success rate, average completion time, or token consumption, our method consistently outperformed the baselines. This indicates that LGC-MARL effectively balances collaboration, efficiency, and resource utilization.

\textbf{Ablation Study.} To delve deeper into the contributions of each module within the LGC-MARL framework, we conducted ablation studies. By removing different modules, we constructed three variants: LGC-MARL (wo critic) without the LLM critic, LGC-MARL (wo reward) without the LLM-generated reward function, and LGC-MARL (wo graph) where the LLM directly interacts with MARL without a graph structure.

As shown in Table \ref{tab:my-table1}, LGC-MARL (wo critic) experienced a significant drop in task success rate due to the LLM planner making more errors without the guidance of the LLM critic. LGC-MARL (wo graph) suffered from increased communication overhead and decreased inter-agent collaboration efficiency due to direct interactions, leading to longer task completion times. LGC-MARL (wo reward) exhibited suboptimal performance across all metrics due to the imperfect reward function, resulting in a less optimized MARL policy.

These experimental results unequivocally demonstrate the effectiveness of each module in LGC-MARL. The LLM critic plays a pivotal role in enhancing decision accuracy, the graph structure effectively reduces communication overhead and improves collaboration efficiency, and the LLM-generated reward function provides high-quality feedback for MARL learning.

\begin{figure}[htbp]
	\centering
		\centering
		\includegraphics[width=0.99\linewidth]{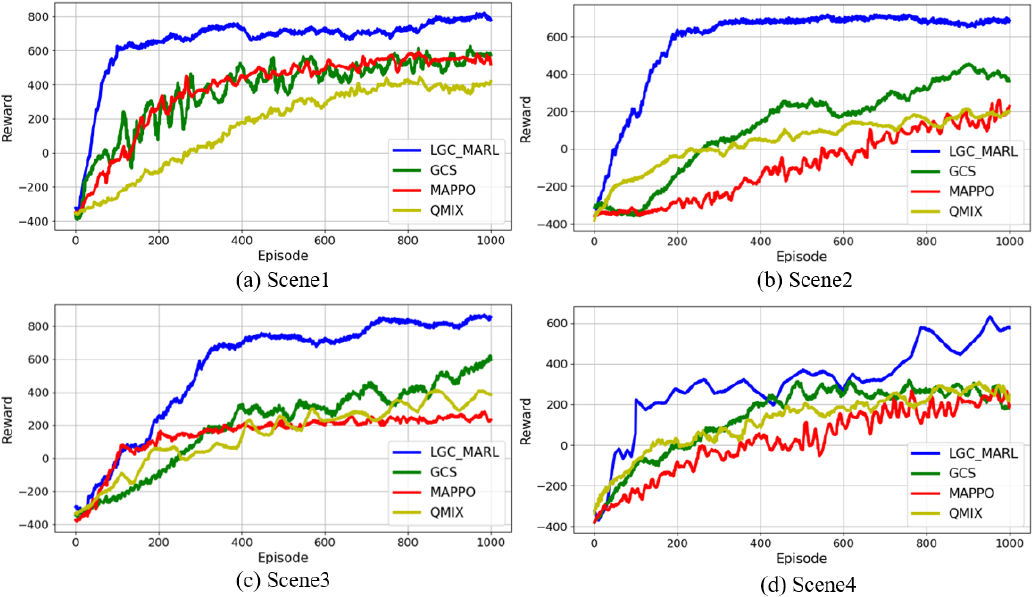}
		\caption{Comparison of different MARL algorithms}
		\label{Comparison of different MARL algorithms}%文中引用该图片代号
\end{figure}

\textbf{Comparison of Different MARL Algorithms.} To comprehensively evaluate the performance of LGC-MARL, we conducted comparative experiments against a suite of state-of-the-art MARL algorithms. Considering that MARL algorithms often struggle to effectively decompose high-level instructions when handling complex tasks, we pre-decomposed complex tasks into subtasks in our comparative experiments to facilitate fair comparisons between different algorithms.

As illustrated in Fig. \ref{Comparison of different MARL algorithms}, our LGC-MARL algorithm consistently outperforms other baseline algorithms across various tasks, achieving significantly higher convergence rewards. In contrast, while classical MARL algorithms like QMIX and MAPPO can enable multi-agent collaboration, they fail to fully consider the interdependence of actions among agents, leading to suboptimal cooperation in complex scenarios and hindering the achievement of higher collaborative rewards. Although GCS \cite{ruan2022gcs} considers the collaborative relationships among agents to some extent, its relatively simple mechanism for generating action collaboration graphs limits its ability to fully leverage environmental information, resulting in slower convergence and less robust learned cooperative policies.
% \vspace{-0.2em}
\begin{figure}[htbp]
	\centering
		\centering
		\includegraphics[width=0.99\linewidth]{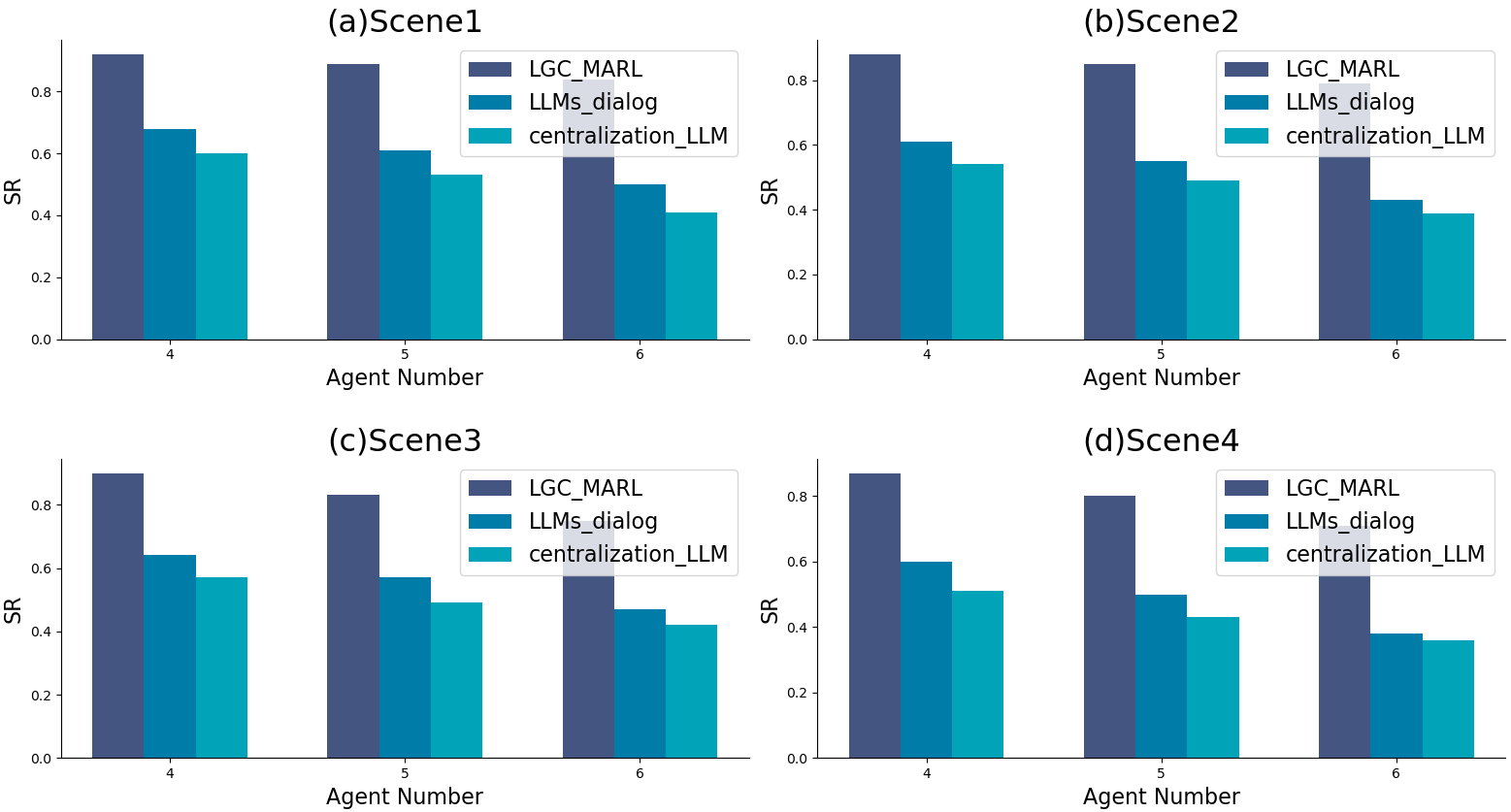}
		\caption{Comparison of different agent numbers}
		\label{Comparison of different agent numbers}%文中引用该图片代号
\end{figure}\textbf{}
% \vspace{-0.2em}

\textbf{Comparison of Different Agent Numbers.} To delve deeper into the scalability of different approaches when handling multi-agent systems, we systematically investigated the impact of the number of agents on algorithm performance. As shown in Fig. 5, as the number of agents increases, both centralized LLM and LLM dialogue methods experience a significant decline in success rate. This is primarily because centralized LLM struggles to effectively manage the complex interactions among a large number of agents, while the LLM dialogue method is prone to lengthy dialogue cycles, leading to inefficient decision-making.

In contrast, our LGC-MARL method exhibits stronger robustness when the number of agents increases, with a smaller decrease in success rate. This indicates that LGC-MARL can effectively handle the complexity of multi-agent systems and maintain high performance even as the number of agents grows. 

\section{CONCLUSIONS}

To enhance the collaboration efficiency and reduce communication overhead in multi-agent systems, we propose LGC-MARL, a framework that integrates MARL with LLMs. The framework consists of two components: an LLM planner and a graph-based collaboration meta policy.
The LLM planner, equipped with rich world knowledge, can decompose complex tasks into a series of executable subtasks and generate an action dependency graph based on the environment. To mitigate the hallucination problem of LLMs and improve task success rates, we equip the LLM planner with an LLM critic to evaluate the rationality of the generated tasks and provide feedback. The graph-based collaboration meta policy guides agents to collaborate based on the action dependency graph and quickly adapts to new task environments through meta-learning. And LLM-based reward function generator generates reward functions tailored to the characteristics of multi-agent systems, ensuring efficient policy learning. Extensive experiments conducted on various scenarios in the AI2-THOR simulation platform demonstrate that LGC-MARL outperforms existing LLM-based and MARL-based methods. While maintaining low token consumption, LGC-MARL consistently achieves high success rates on tasks.
\section{Acknowledgements}
This paper is supported by the Key Research and Development Program of Guangdong Province under grant No.2021B0101400003. Corresponding author is Jianzong Wang from Ping An Technology (Shenzhen) Co., Ltd.


%%%%%%%%%%%%%%%%%%%%%%%%%%%%%%%%%%%%%%%%%%%%%%%%%%%%%%%%%%%%%%%%%%%%%%%%%%%%%%%%

%%%%%%%%%%%%%%%%%%%%%%%%%%%%%%%%%%%%%%%%%%%%%%%%%%%%%%%%%%%%%%%%%%%%%%%%%%%%%%%%



% \section*{ACKNOWLEDGMENT}

% The preferred spelling of the word ÒacknowledgmentÓ in America is without an ÒeÓ after the ÒgÓ. Avoid the stilted expression, ÒOne of us (R. B. G.) thanks . . .Ó  Instead, try ÒR. B. G. thanksÓ. Put sponsor acknowledgments in the unnumbered footnote on the first page.

% %%%%%%%%%%%%%%%%%%%%%%%%%%%%%%%%%%%%%%%%%%%%%%%%%%%%%%%%%%%%%%%%%%%%%%%%%%%%%%%%

% References are important to the reader; therefore, each citation must be complete and correct. If at all possible, references should be commonly available publications.

\bibliographystyle{IEEEtran}
\bibliography{IEEEabrv}

\begin{thebibliography}{10}
\providecommand{\url}[1]{#1}
\csname url@rmstyle\endcsname
\providecommand{\newblock}{\relax}
\providecommand{\bibinfo}[2]{#2}
\providecommand\BIBentrySTDinterwordspacing{\spaceskip=0pt\relax}
\providecommand\BIBentryALTinterwordstretchfactor{4}
\providecommand\BIBentryALTinterwordspacing{\spaceskip=\fontdimen2\font plus
\BIBentryALTinterwordstretchfactor\fontdimen3\font minus \fontdimen4\font\relax}
\providecommand\BIBforeignlanguage[2]{{%
\expandafter\ifx\csname l@#1\endcsname\relax
\typeout{** WARNING: IEEEtran.bst: No hyphenation pattern has been}%
\typeout{** loaded for the language `#1'. Using the pattern for}%
\typeout{** the default language instead.}%
\else
\language=\csname l@#1\endcsname
\fi
#2}}

\bibitem{venkata2023kt}
S.~S.~O. Venkata, R.~Parasuraman, and R.~Pidaparti, ``Kt-bt: A framework for knowledge transfer through behavior trees in multirobot systems,'' \emph{IEEE Transactions on Robotics}, 2023.

\bibitem{liu2023robust}
W.~Liu, K.~Leahy, Z.~Serlin, and C.~Belta, ``Robust multi-agent coordination from catl+ specifications,'' in \emph{2023 American Control Conference (ACC)}.\hskip 1em plus 0.5em minus 0.4em\relax IEEE, 2023, pp. 3529--3534.

\bibitem{martin2023multi}
J.~G. Martin, F.~J. Muros, J.~M. Maestre, and E.~F. Camacho, ``Multi-robot task allocation clustering based on game theory,'' \emph{Robotics and Autonomous Systems}, vol. 161, p. 104314, 2023.

\bibitem{chakraa2023optimization}
H.~Chakraa, F.~Gu{\'e}rin, E.~Leclercq, and D.~Lefebvre, ``Optimization techniques for multi-robot task allocation problems: Review on the state-of-the-art,'' \emph{Robotics and Autonomous Systems}, p. 104492, 2023.

\bibitem{chen2024scalable}
Y.~Chen, J.~Arkin, Y.~Zhang, N.~Roy, and C.~Fan, ``Scalable multi-robot collaboration with large language models: Centralized or decentralized systems?'' in \emph{2024 IEEE International Conference on Robotics and Automation (ICRA)}.\hskip 1em plus 0.5em minus 0.4em\relax IEEE, 2024, pp. 4311--4317.

\bibitem{zhang2022multi}
H.~Zhang, J.~Chen, J.~Li, B.~Williams, and S.~Koenig, ``Multi-agent path finding for precedence-constrained goal sequences,'' in \emph{International Joint Conference on Autonomous Agents and Multiagent Systems (AAMAS)}, 2022.

\bibitem{chen2024meta}
J.~Chen, Y.~Gao, J.~Hu, F.~Deng, and T.~L. Lam, ``Meta-reinforcement learning based cooperative surface inspection of 3d uncertain structures using multi-robot systems,'' in \emph{2024 IEEE International Conference on Robotics and Automation (ICRA)}.\hskip 1em plus 0.5em minus 0.4em\relax IEEE, 2024, pp. 7201--7207.

\bibitem{yao2020multi}
S.~Yao, G.~Chen, L.~Pan, J.~Ma, J.~Ji, and X.~Chen, ``Multi-robot collision avoidance with map-based deep reinforcement learning,'' in \emph{2020 IEEE 32nd International Conference on Tools with Artificial Intelligence (ICTAI)}.\hskip 1em plus 0.5em minus 0.4em\relax IEEE, 2020, pp. 532--539.

\bibitem{zhang2021multi}
K.~Zhang, Z.~Yang, and T.~Ba{\c{s}}ar, ``Multi-agent reinforcement learning: A selective overview of theories and algorithms,'' \emph{Handbook of reinforcement learning and control}, pp. 321--384, 2021.

\bibitem{oroojlooy2023review}
A.~Oroojlooy and D.~Hajinezhad, ``A review of cooperative multi-agent deep reinforcement learning,'' \emph{Applied Intelligence}, vol.~53, no.~11, pp. 13\,677--13\,722, 2023.

\bibitem{kang2023cooperative}
H.~Kang, X.~Chang, J.~Mi{\v{s}}i{\'c}, V.~B. Mi{\v{s}}i{\'c}, J.~Fan, and Y.~Liu, ``Cooperative uav resource allocation and task offloading in hierarchical aerial computing systems: A mappo-based approach,'' \emph{IEEE Internet of Things Journal}, vol.~10, no.~12, pp. 10\,497--10\,509, 2023.

\bibitem{rashid2020weighted}
T.~Rashid, G.~Farquhar, B.~Peng, and S.~Whiteson, ``Weighted qmix: Expanding monotonic value function factorisation for deep multi-agent reinforcement learning,'' \emph{Advances in Neural Information Processing Systems (NeurIPS)}, vol.~33, pp. 10\,199--10\,210, 2020.

\bibitem{huang2022inner}
W.~Huang, F.~Xia, T.~Xiao, H.~Chan, J.~Liang, P.~Florence, A.~Zeng, J.~Tompson, I.~Mordatch, Y.~Chebotar, \emph{et~al.}, ``Inner monologue: Embodied reasoning through planning with language models,'' \emph{Conference on Robot Learning (CoRL)}, 2022.

\bibitem{liang2023code}
J.~Liang, W.~Huang, F.~Xia, P.~Xu, K.~Hausman, B.~Ichter, P.~Florence, and A.~Zeng, ``Code as policies: Language model programs for embodied control,'' in \emph{2023 IEEE International Conference on Robotics and Automation (ICRA)}.\hskip 1em plus 0.5em minus 0.4em\relax IEEE, 2023, pp. 9493--9500.

\bibitem{singh2023progprompt}
I.~Singh, V.~Blukis, A.~Mousavian, A.~Goyal, D.~Xu, J.~Tremblay, D.~Fox, J.~Thomason, and A.~Garg, ``Progprompt: Generating situated robot task plans using large language models,'' in \emph{2023 IEEE International Conference on Robotics and Automation (ICRA)}.\hskip 1em plus 0.5em minus 0.4em\relax IEEE, 2023, pp. 11\,523--11\,530.

\bibitem{wu2023tidybot}
J.~Wu, R.~Antonova, A.~Kan, M.~Lepert, A.~Zeng, S.~Song, J.~Bohg, S.~Rusinkiewicz, and T.~Funkhouser, ``Tidybot: Personalized robot assistance with large language models,'' \emph{Autonomous Robots}, vol.~47, no.~8, pp. 1087--1102, 2023.

\bibitem{lin2023text2motion}
K.~Lin, C.~Agia, T.~Migimatsu, M.~Pavone, and J.~Bohg, ``Text2motion: From natural language instructions to feasible plans,'' \emph{Autonomous Robots}, vol.~47, no.~8, pp. 1345--1365, 2023.

\bibitem{ding2023task}
Y.~Ding, X.~Zhang, C.~Paxton, and S.~Zhang, ``Task and motion planning with large language models for object rearrangement,'' in \emph{2023 IEEE/RSJ International Conference on Intelligent Robots and Systems (IROS)}.\hskip 1em plus 0.5em minus 0.4em\relax IEEE, 2023, pp. 2086--2092.

\bibitem{DBLP:conf/corl/0003GFKLACEHHIX23}
W.~Yu, N.~Gileadi, C.~Fu, S.~Kirmani, K.~Lee, M.~G. Arenas, H.~L. Chiang, T.~Erez, L.~Hasenclever, J.~Humplik, B.~Ichter, T.~Xiao, P.~Xu, A.~Zeng, T.~Zhang, N.~Heess, D.~Sadigh, J.~Tan, Y.~Tassa, and F.~Xia, ``Language to rewards for robotic skill synthesis,'' in \emph{Conference on Robot Learning (CoRL)}, 2023.

\bibitem{chen2024autotamp}
Y.~Chen, J.~Arkin, C.~Dawson, Y.~Zhang, N.~Roy, and C.~Fan, ``Autotamp: Autoregressive task and motion planning with llms as translators and checkers,'' in \emph{2024 IEEE International Conference on Robotics and Automation (ICRA)}.\hskip 1em plus 0.5em minus 0.4em\relax IEEE, 2024, pp. 6695--6702.

\bibitem{irpan2022can}
A.~Irpan, A.~Herzog, A.~T. Toshev, A.~Zeng, A.~Brohan, B.~A. Ichter, B.~David, C.~Parada, C.~Finn, C.~Tan, \emph{et~al.}, ``Do as i can, not as i say: Grounding language in robotic affordances,'' in \emph{Conference on Robot Learning (CoRL)}, 2022.

\bibitem{chen2024llm}
R.~Chen, W.~Song, W.~Zu, Z.~Dong, Z.~Guo, F.~Sun, Z.~Tian, and J.~Wang, ``An llm-driven framework for multiple-vehicle dispatching and navigation in smart city landscapes,'' in \emph{2024 IEEE International Conference on Robotics and Automation (ICRA)}.\hskip 1em plus 0.5em minus 0.4em\relax IEEE, 2024, pp. 2147--2153.

\bibitem{mandi2024roco}
Z.~Mandi, S.~Jain, and S.~Song, ``Roco: Dialectic multi-robot collaboration with large language models,'' in \emph{2024 IEEE International Conference on Robotics and Automation (ICRA)}.\hskip 1em plus 0.5em minus 0.4em\relax IEEE, 2024, pp. 286--299.

\bibitem{liu2023dynamic}
Z.~Liu, Y.~Zhang, P.~Li, Y.~Liu, and D.~Yang, ``Dynamic llm-agent network: An llm-agent collaboration framework with agent team optimization,'' \emph{arXiv preprint arXiv:2310.02170}, 2023.

\bibitem{DBLP:conf/cvpr/Yang0LTLS00S24}
Y.~Yang, T.~Zhou, K.~Li, D.~Tao, L.~Li, L.~Shen, X.~He, J.~Jiang, and Y.~Shi, ``Embodied multi-modal agent trained by an {LLM} from a parallel textworld,'' in \emph{{IEEE/CVF} Conference on Computer Vision and Pattern Recognition (CVPR)}.\hskip 1em plus 0.5em minus 0.4em\relax {IEEE}, 2024, pp. 26\,265--26\,275.

\bibitem{chen2023multi}
H.~Chen, W.~Ji, L.~Xu, and S.~Zhao, ``Multi-agent consensus seeking via large language models,'' \emph{arXiv preprint arXiv:2310.20151}, 2023.

\bibitem{zhang2023building}
H.~Zhang, W.~Du, J.~Shan, Q.~Zhou, Y.~Du, J.~B. Tenenbaum, T.~Shu, and C.~Gan, ``Building cooperative embodied agents modularly with large language models,'' \emph{International Conference on Learning Representations (ICLR)}, 2024.

\bibitem{DBLP:conf/iros/KannanVM24}
S.~S. Kannan, V.~L.~N. Venkatesh, and B.~Min, ``{SMART-LLM:} smart multi-agent robot task planning using large language models,'' in \emph{2024 {IEEE/RSJ} International Conference on Intelligent Robots and Systems (IROS)}.\hskip 1em plus 0.5em minus 0.4em\relax {IEEE}, 2024, pp. 12\,140--12\,147.

\bibitem{DBLP:conf/iclr/SzotSAMMTMHT24}
A.~Szot, M.~Schwarzer, H.~Agrawal, B.~Mazoure, R.~Metcalf, W.~Talbott, N.~Mackraz, R.~D. Hjelm, and A.~T. Toshev, ``Large language models as generalizable policies for embodied tasks,'' in \emph{International Conference on Learning Representations (ICLR)}, 2024.

\bibitem{wu2024spring}
Y.~Wu, S.~Y. Min, S.~Prabhumoye, Y.~Bisk, R.~R. Salakhutdinov, A.~Azaria, T.~M. Mitchell, and Y.~Li, ``Spring: Studying papers and reasoning to play games,'' \emph{Advances in Neural Information Processing Systems (NeurIPS)}, vol.~36, 2024.

\bibitem{carta2023grounding}
T.~Carta, C.~Romac, T.~Wolf, S.~Lamprier, O.~Sigaud, and P.-Y. Oudeyer, ``Grounding large language models in interactive environments with online reinforcement learning,'' in \emph{International Conference on Machine Learning (ICML)}.\hskip 1em plus 0.5em minus 0.4em\relax PMLR, 2023, pp. 3676--3713.

\bibitem{baai2023plan4mc}
P.~BAAI, ``Plan4mc: Skill reinforcement learning and planning for open-world minecraft tasks,'' \emph{arXiv preprint arXiv:2303.16563}, 2023.

\bibitem{ma2023eureka}
Y.~J. Ma, W.~Liang, G.~Wang, D.-A. Huang, O.~Bastani, D.~Jayaraman, Y.~Zhu, L.~Fan, and A.~Anandkumar, ``Eureka: Human-level reward design via coding large language models,'' \emph{International Conference on Learning Representations (ICLR)}, 2024.

\bibitem{xie2023text2reward}
T.~Xie, S.~Zhao, C.~H. Wu, Y.~Liu, Q.~Luo, V.~Zhong, Y.~Yang, and T.~Yu, ``Text2reward: Automated dense reward function generation for reinforcement learning,'' \emph{International Conference on Learning Representations (ICLR)}, 2024.

\bibitem{kolve2017ai2}
E.~Kolve, R.~Mottaghi, W.~Han, E.~VanderBilt, L.~Weihs, A.~Herrasti, M.~Deitke, K.~Ehsani, D.~Gordon, Y.~Zhu, \emph{et~al.}, ``Ai2-thor: An interactive 3d environment for visual ai,'' \emph{arXiv preprint arXiv:1712.05474}, 2017.

\bibitem{ruan2022gcs}
J.~Ruan, Y.~Du, X.~Xiong, D.~Xing, X.~Li, L.~Meng, H.~Zhang, J.~Wang, and B.~Xu, ``Gcs: Graph-based coordination strategy for multi-agent reinforcement learning,'' \emph{International Joint Conference on Autonomous Agents and Multiagent Systems (AAMAS)}, 2022.

\end{thebibliography}
\end{document}